%% file: main.tex
\documentclass[sigconf]{acmart}
\AtBeginDocument{%
  }

\PassOptionsToPackage{prologue,dvipsnames}{xcolor}
\usepackage{xspace}
\usepackage{bbm}
\usepackage{caption}
\usepackage{wrapfig}
\usepackage{adjustbox}
\usepackage{subcaption}

\usepackage{multirow}

\newcommand{\eg}[1]{\emph{e.g.}}

\newcommand{\method}{MarkSplatter\xspace}

\newcommand{\Tref}[1]{Table~\ref{#1}}

\newcommand{\Fref}[1]{Figure~\ref{#1}}

\setcopyright{acmlicensed}
\copyrightyear{2025}
\acmYear{2025}
\setcopyright{cc}
\setcctype{by}
\acmConference[MM '25]{Proceedings of the 33rd ACM International Conference
on Multimedia}{October 27--31, 2025}{Dublin, Ireland}
\acmBooktitle{Proceedings of the 33rd ACM International Conference on
Multimedia (MM '25), October 27--31, 2025, Dublin,
Ireland}
\acmDOI{10.1145/3746027.3758144}
\acmISBN{979-8-4007-2035-2/2025/10}

\acmSubmissionID{15}

\begin{document}

%%
%% The "title" command has an optional parameter,xiufwe
%% allowing the author to define a "short title" to be used in page headers.
% \title{MASplatt3R: Lifting MASt3R to Multi-View 3D Gaussian Recontruction Model for 3D Content Generation}
\title{MarkSplatter: Generalizable Watermarking for 3D Gaussian Splatting Model via Splatter Image Structure}

%%
%% The "author" command and its associated commands are used to define
%% the authors and their affiliations.
%% Of note is the shared affiliation of the first two authors, and the
%% "authornote" and "authornotemark" commands
%% used to denote shared contribution to the research.
\author{Xiufeng Huang}
\email{xiufenghuang@life.hkbu.edu.hk}
\affiliation{%
  \institution{Department of Computer Science, Hong Kong Baptist University}
  \city{Hong Kong}
  % \state{Hong Kong SAR}
  \country{China}
}

\author{Ziyuan Luo}
\email{ziyuanluo@life.hkbu.edu.hk}
\affiliation{%
  \institution{Department of Computer Science, Hong Kong Baptist University}
  \city{Hong Kong}
  % \state{Hong Kong SAR}
  \country{China}
}

\author{Qi Song}
\email{qisong@life.hkbu.edu.hk}
\affiliation{%
  \institution{Department of Computer Science, Hong Kong Baptist University}
  \city{Hong Kong}
  % \state{Hong Kong SAR}
  \country{China}
}

\author{Ruofei Wang}
\email{ruofei@life.hkbu.edu.hk}
\affiliation{%
  \institution{Department of Computer Science, Hong Kong Baptist University}
  \city{Hong Kong}
  % \state{Hong Kong SAR}
  \country{China}
}

\author{Renjie Wan}
\email{renjiewan@hkbu.edu.hk}
\authornote{Corresponding author.}
\affiliation{%
  \institution{Department of Computer Science, Hong Kong Baptist University}
  \city{Hong Kong}
  % \state{Hong Kong SAR}
  \country{China}
}

%%
%% By default, the full list of authors will be used in the page
%% headers. Often, this list is too long, and will overlap
%% other information printed in the page headers. This command allows
%% the author to define a more concise list
%% of authors' names for this purpose.
\renewcommand{\shortauthors}{Xiufeng Huang, Ziyuan Luo, Qi Song, Ruofei Wang, and Renjie Wan}

%%
%% The abstract is a short summary of the work to be presented in the
%% article.
% \begin{abstract}
%   A clear and well-documented \LaTeX\ document is presented as an
%   article formatted for publication by ACM in a conference proceedings
%   or journal publication. Based on the ``acmart'' document class, this
%   article presents and explains many of the common variations, as well
%   as many of the formatting elements an author may use in the
%   preparation of the documentation of their work.
% \end{abstract}
\input{sec/0_abstract}

%%
%% The code below is generated by the tool at http://dl.acm.org/ccs.cfm.
%% Please copy and paste the code instead of the example below.
%%
% \begin{CCSXML}
% <ccs2012>
%  <concept>
%   <concept_id>00000000.0000000.0000000</concept_id>
%   <concept_desc>Do Not Use This Code, Generate the Correct Terms for Your Paper</concept_desc>
%   <concept_significance>500</concept_significance>
%  </concept>
%  <concept>
%   <concept_id>00000000.00000000.00000000</concept_id>
%   <concept_desc>Do Not Use This Code, Generate the Correct Terms for Your Paper</concept_desc>
%   <concept_significance>300</concept_significance>
%  </concept>
%  <concept>
%   <concept_id>00000000.00000000.00000000</concept_id>
%   <concept_desc>Do Not Use This Code, Generate the Correct Terms for Your Paper</concept_desc>
%   <concept_significance>100</concept_significance>
%  </concept>
%  <concept>
%   <concept_id>00000000.00000000.00000000</concept_id>
%   <concept_desc>Do Not Use This Code, Generate the Correct Terms for Your Paper</concept_desc>
%   <concept_significance>100</concept_significance>
%  </concept>
% </ccs2012>
% \end{CCSXML}
% \ccsdesc[500]{Information systems~Multimedia content creation}



% \ccsdesc[300]{Do Not Use This Code~Generate the Correct Terms for Your Paper}
% \ccsdesc{Do Not Use This Code~Generate the Correct Terms for Your Paper}
% \ccsdesc[100]{Do Not Use This Code~Generate the Correct Terms for Your Paper}

%%
%% Keywords. The author(s) should pick words that accurately describe
%% the work being presented. Separate the keywords with commas.
%% A "teaser" image appears between the author and affiliation
%% information and the body of the document, and typically spans the
%% page.

% \received{20 February 2007}
% \received[revised]{12 March 2009}
% \received[accepted]{5 June 2009}

%%
%% This command processes the author and affiliation and title
%% information and builds the first part of the formatted document.
\maketitle

\input{sec/1_intro}

\input{sec/2_related_work}

\input{sec/3_preliminary}
\input{sec/4_proposed_method}

\input{sec/5_experiments}

\input{sec/6_conclusion}

\input{sec/7_acknowledgement}

%%
%% The next two lines define the bibliography style to be used, and
%% the bibliography file.
\bibliographystyle{ACM-Reference-Format}
\bibliography{main}

%%
%% If your work has an appendix, this is the place to put it.
% \appendix
% \input{sec/X_suppl}

\end{document}

%% file: sec/0_abstract.tex
\begin{abstract}
The growing popularity of 3D Gaussian Splatting (3DGS) has intensified the need for effective copyright protection. Current 3DGS watermarking methods rely on computationally expensive fine-tuning procedures for each predefined message.
We propose the first generalizable watermarking framework that enables efficient protection of Splatter Image-based 3DGS models through a single forward pass. We introduce GaussianBridge that transforms unstructured 3D Gaussians into Splatter Image format, enabling direct neural processing for arbitrary message embedding. 
To ensure imperceptibility, we design a Gaussian-Uncertainty-Perceptual heatmap prediction strategy for preserving visual quality. For robust message recovery, we develop a dense segmentation-based extraction mechanism that maintains reliable extraction even when watermarked objects occupy minimal regions in rendered views.
Project page: https://kevinhuangxf.github.io/marksplatter.
% Extensive experiments demonstrate that our method achieves state-of-the-art performance in both reconstruction quality and watermark robustness.
\end{abstract}

% while existing watermark extractors struggle with reliability when watermarked content appears at varying scales. 

%% file: sec/1_intro.tex
\begin{figure}[t]
  \centering
  % \fbox{\rule{0pt}{2in} \rule{0.9\linewidth}{0pt}}
  \includegraphics[width=\linewidth]{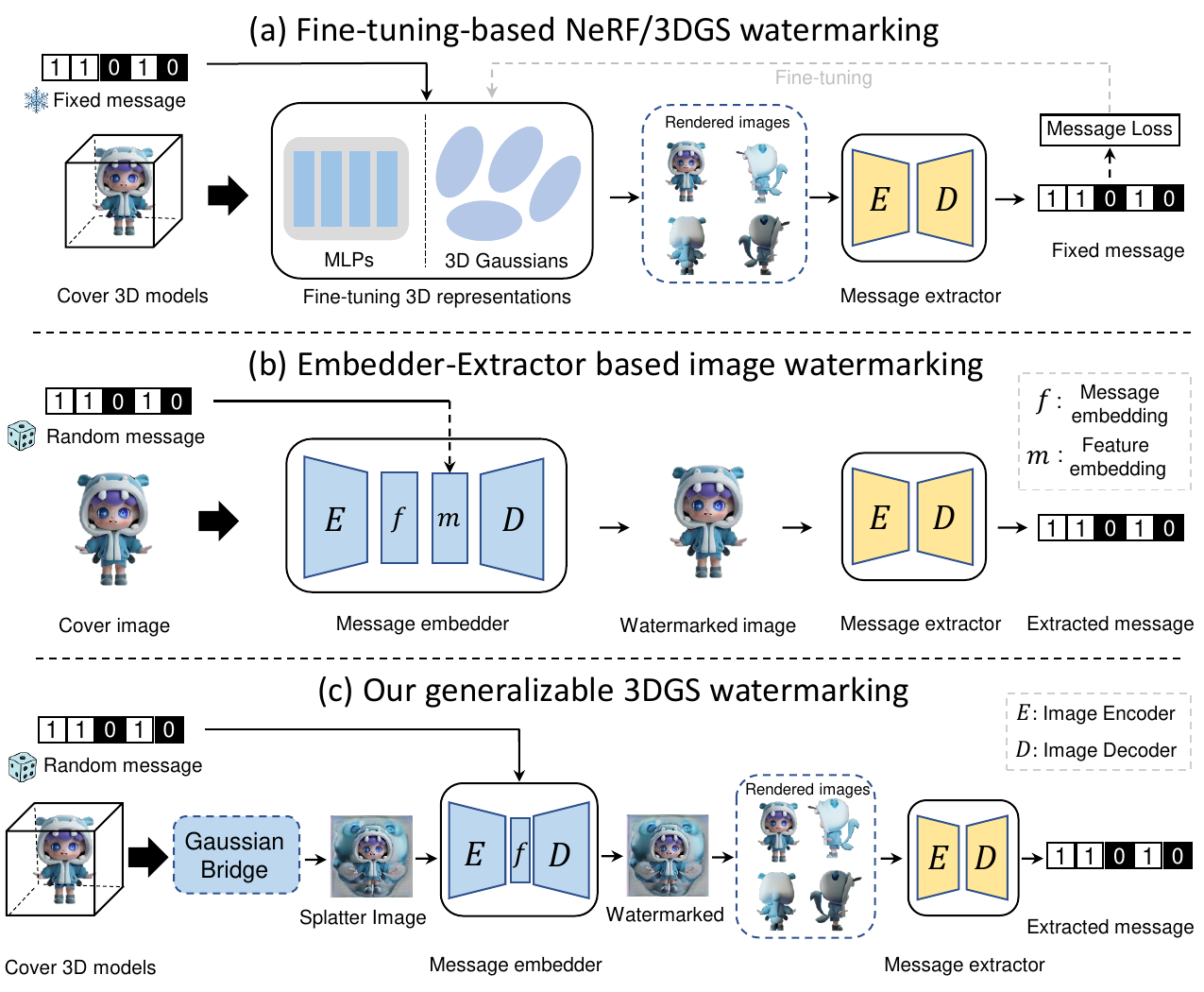}

   % \caption{Color-editing on watermarking embedded NeRF models. CopyRNeRF and StegaNeRF fail to recover the secrete message. Our method can reliably retrieve the secrete message even with various color-editing.}
   \caption{Comparison of different watermarking approaches. 
   (a) Current fine-tuning-based NeRF/3DGS watermarking requires model-specific optimization. (b) Generalizable image watermarking using an embedder-extractor architecture for arbitrary images. (c) Our proposed generalizable 3DGS watermarking framework enables efficient random message embedding for Splatter Image-based 3DGS. By using GaussianBridge as an optional plug-in, our framework can be extended to point cloud-based 3DGS.
   }

   \label{fig:proposed_scene}
\end{figure}

% \vspace{-20pt}

\section{Introduction}
\label{sec:introduction}

%\textcolor{brown}{
%The recent great progress on the generative models for 3D Gaussians Splatting (3DGS) \cite{hong2023lrm, szymanowicz2024splatter, zou2024triplane, tang2024lgm} had made the protection of copyright for these 3D digital assets more critical than ever.
%However, current 3DGS watermarking methods \cite{huang2024geometrysticker, huang2024gaussianmarker} rely on fine-tuning the 3DGS parameters with a fixed input message, which is time-consuming and impractical for real-world applications.
%Regarding the success of image watermarking in protecting 2D media \cite{zhu2018hidden, jia2021mbrs, sander2025watermark}, 
%\textit{how can we protect the 3DGS models in a generative and effective way just like image watermarking?}
%}

The remarkable advancements in 3D Gaussian Splatting generative models \cite{hong2023lrm, szymanowicz2024splatter, tang2024lgm} have intensified the need for effective copyright protection of 3D assets. 
However, as we discussed in \textbf{\Fref{fig:proposed_scene} (a)}, in current watermarking settings for 3DGS \cite{huang2024geometrysticker, huang2024gaussianmarker}, existing methods rely on computationally expensive fine-tuning procedures for each predefined message. 
Such a per-scene-specific setting can only support fixed watermark embedding.
Once the watermark is embedded, it cannot be dynamically modified without repeating the entire fine-tuning process, severely limiting practical deployment in real-world applications.

A more practical scenario should be a generalizable watermarking pipeline that enables on-demand watermark embedding into pre-trained 3DGS models without requiring repeated optimization procedures. 
This framework would allow for seamless embedding of arbitrary messages through a single forward pass, facilitating efficient adaptation to different copyright messages and 3D contents while preserving rendering quality and maintaining robustness.

The effectiveness of such generalizable frameworks has been well-established in image watermarking \cite{zhu2018hidden, jia2021mbrs, sander2025watermark} as shown in \textbf{\Fref{fig:proposed_scene} (b)}. 
In current settings for image watermarking, neural networks process images to embed and extract watermarks in a single forward pass, enabling rapid implementation of copyright protection across different contents. 
However, directly adapting such approaches for 3DGS models encounters significant challenges. \textbf{First}, the absence of a framework that can directly process 3DGS representations limits the development of efficient watermarking solutions. Unlike 2D images with regular grid structures, 3DGS models consist of unstructured point cloud-based 3D Gaussians with varying attributes, making it non-trivial to apply existing neural network architectures for watermark embedding and extraction. \textbf{Second}, existing image watermark extractors~\cite{zhu2018hidden, jia2021mbrs} demonstrate limited effectiveness when processing rendered images from 3D models. This limitation is particularly evident when watermarked content only occupies a small portion of the rendered image due to camera zoom-out or perspective changes, a common scenario in 3D visualization. Traditional extractors~\cite{zhu2018hidden, jia2021mbrs}, which decode watermarks globally from the entire image, often fail to accurately extract messages when the watermarked object appears at different scales and positions. 
Without addressing these challenges, current watermarking methods rely on computationally intensive optimization procedures that lack the flexibility and efficiency needed for practical deployment.

To address these challenges, we propose a novel framework for generalizable watermarking of 3D Gaussian Splatting models~\cite{kerbl3Dgaussians} as illustrated in \textbf{\Fref{fig:proposed_scene} (c)}. We identify a new structure for 3DGS, aka the Splatter Image~\cite{szymanowicz2024splatter}. It is a grid-based 3DGS format, which maintains the same information from the point cloud-based 3DGS model. 
% It can also be easily transferred to the classical point cloud structure. 
By flattening the 2D grid-based 3DGS data into the 1D unstructured format, the Splatter Image representation can be seamlessly converted to the native point cloud-based 3DGS structure.
More importantly, such a format is compatible with conventional neural networks, which enables direct processing through standard neural architectures~\cite{he2016deep, dosovitskiy2020vit}. The data owners can directly utilize trained neural networks for message embedding and extraction, just like the established convenient settings in image watermarking~\cite{zhu2018hidden, jia2021mbrs}. 

% This conversion enables seamless integration with existing image watermarking frameworks, allowing arbitrary messages to be embedded via a single forward pass while maintaining high-fidelity 3D reconstruction quality for downstream tasks.
% It bridges the gap between unstructured 3D Gaussian representations and 2D image-based watermarking workflows by enabling seamless integration between these domains.
% \renjie{For the encoder-decoder structure used for the processing of splatter images, }
 
% designed for generalizable and robust 3DGS watermarking. 
% To make 3DGS models compatible with existing image foundation models, We treat 3DGS models as multi-view Splatter Images \cite{szymanowicz2024splatter}, which can organize the 3DGS models from an unstructured point cloud format into a spatially structured image format.Based on the acquired Splatter Images, To achieve visual quality preservation, we further propose a novel Just-Noticeable-Difference (JND) heatmap prediction for Splatter Images to make the embedded watermark \textbf{imperceptible}.

We design an \textbf{embedder-extractor} watermarking framework for 3DGS message embedding and extraction, similar to the proven convenient setup in image watermarking~\cite{zhu2018hidden, jang2024waterf}. 
Our message embedder utilizes an encoder-decoder architecture that jointly optimizes two complementary mechanisms: in-attribute embedding for encoding messages within Splatter Image's color attributes, and cross-attribute constraints to regulate perturbation magnitudes. 
These perturbations are adaptively modulated through a Gaussian Uncertainty Perceptual (GUP) heatmap, which constrains attributes with high perceptual sensitivity to make the embedded watermarks \textbf{imperceptible}.

For message extraction, we propose a dense segmentation approach that treats watermark extraction as a pixel-wise message localization and prediction task, ensuring \textbf{robust} extraction even when watermarked regions occupy minimal image space. This segmentation-based mechanism overcomes the limitations of traditional extractors that rely on the global watermark embedding and maintains effectiveness across various viewing conditions. 

With many point cloud-based 3DGS models already in existence, we introduce GaussianBridge, a plug-and-play module that transforms unstructured point cloud-based 3DGS models into Splatter Image format. 
Leveraging a Gaussian reconstruction module as the core, we devise a method to fully capture point cloud information. 
This promotes the protection for both Splatter Image and point cloud-based 3DGS models.
Upon completion of the watermarking procedure on Splatter Image, the watermarked 3DGS models can also be easily converted back to the point cloud structure by the flattening operation.

In summary, our contributions are listed below:
\begin{itemize}
  \item We propose the first generalizable 3DGS watermarking framework that operates directly on 3D Gaussians via the Splatter Image structure, eliminating the need for computationally expensive per-scene optimization while enabling scalable protection across diverse 3DGS models.
  % \item We present GaussianBridge, which seamlessly converts 3DGS models from the point cloud format to the Splatter Image format, enabling generalized protection for existing point cloud-based 3DGS models.
  \item We introduce GaussianBridge, a novel conversion module that seamlessly transforms 3DGS models from their native point cloud format to the Splatter Image representation, enabling unified watermarking protection for existing point cloud-based 3DGS architectures.
  % \item We introduce a novel embedder-extractor structure to bridge the 2D and 3D watermarking gap for message embedding and extraction, based on the Splatter Image format and GaussianBridge module.
  \item We introduce a novel embedder-extractor architecture that achieves the first unified neural network framework for both image and 3DGS watermarking, effectively addressing the domain gap between 2D and 3D watermarking via the Splatter Image structure and GaussianBridge module.
\end{itemize}
We call our solution \textbf{\method}. With \method, data owners can apply our generalizable 3DGS watermarking to protect newly created Splatter Image-based 3DGS models~\cite{tang2024lgm, xu2024grm, huang2025stereo} directly. 
They can also effortlessly transform point cloud-based 3DGS models into the Splatter Image structure. \method~enhances the security of 3DGS model distribution.
Extensive experiments demonstrate that our method achieves advanced performance in both reconstruction quality and watermark robustness.

%% file: sec/2_related_work.tex
\section{Related work}
\label{sec:related_word}

% \paragraph{3D Gaussian Splatting.}
\noindent\textbf{3D Gaussian Splatting.}
3DGS \cite{kerbl3Dgaussians} has been rapidly adopted across multiple domains and demonstrated remarkable 3D reconstruction results.
There is a growing interest in reconstructing 3D objects from sparse inputs for generative tasks \cite{guo2024depth, ni2024colnerf, yu2021pixelnerf, li2024variational}, following two main approaches: per-scene optimization methods and feed-forward inference methods. 
The former methods typically leverage multi-view geometry constraints to jointly optimize rendering results and camera poses \cite{niemeyer2022regnerf, truong2023sparf, deng2022depth}.
Feed-forward inference methods \cite{yu2021pixelnerf, wang2021ibrnet, chen2024mvsplat} can reconstruct entire scenes in a single pass without additional optimization.
Splatter Image \cite{szymanowicz2024splatter} is the first method to reconstruct entire 3DGS model from just one input image, which can convert 3DGS models from unstructured point-cloud format into a structured image-like representation. Recent methods like LGM \cite{tang2024lgm}, GRM \cite{xu2024grm}, and GS-LRM \cite{gslrm2024} have further developed this approach using various architectures to reconstruct Gaussian features, making them compatible with image foundation models \cite{dosovitskiy2020vit}.

% LGM \cite{tang2024lgm} utilizes the UNet model to reconstruct multi-view Gaussian Features to represent 3DGS models.
% Similarly, GRM \cite{xu2024grm} and GS-LRM \cite{gslrm2024} propose transformer-based architecture to reconstruct multi-view input into per-pixel Gaussians.
% Both the multi-view Gaussian features from LGM \cite{tang2024lgm} and per-pixel Gaussians from GRM \cite{xu2024grm} and GS-LRM \cite{gslrm2024} can be regarded as the Splatter Image representations, which is compatible with the image foundation models \cite{dosovitskiy2020vit}.

\noindent\textbf{Image watermarking.}
Traditional image watermarking approaches embed information either in spatial domains \cite{van1994digital} or frequency domains using techniques such as Discrete Wavelet Transform (DWT) and Singular Value Decomposition (SVD) \cite{Lai_Tsai_2010, pardhu2016digital}.
The advent of deep learning has led to significant advances in image watermarking, achieving both high quality and robustness \cite{tancik2020stegastamp, weng2019high, wengrowski2019light, yang2021robust, zhang2020udh, zhang2019robust, wang2024spy}.
HiDDeN \cite{zhu2018hidden}, a pioneering deep learning-based method, demonstrated superior performance compared to traditional approaches. Subsequent research has enhanced deep learning-based watermarking through various techniques, such as scalable residual connections~\cite{ahmadi2020redmark}, attention mechanisms~\cite{zhang2020robust, yu2020attention}, and invertible networks \cite{ma2022towards, fang2023flow}. Recently, the Watermark Anything Model (WAM)~\cite{sander2025watermark} reformulates watermark extraction as a segmentation task, enabling pixel-level detection and reliable message extraction from partial image regions. This makes WAM especially effective for images with white or transparent backgrounds, such as rendered 3D objects.

\noindent\textbf{3D watermarking.}
3D watermarking can differ significantly from image watermarking.
Traditional 3D watermarking approaches are designed for explicit 3D models \cite{praun1999robust, wu20153d, chen2023mimic3d, son2017perceptual, yoo2022deep}.
Recent works \cite{luo2023copyrnerf, song2024protecting, huang2024geometrysticker, song2024geometry, luo2025nerf, luomantlemark} have shown great progress on neural rendering-based 3D models, including Neural Radience Fields (NeRF) and 3DGS.
CopyRNeRF~\cite{luo2023copyrnerf} introduces watermarked color representations for invisible NeRF protection with high rendering quality. 
WateRF~\cite{jang2024waterf} employs DWT-based watermarking in NeRF space with patch-wise optimization for better robustness.
GaussianMarker~\cite{huang2024gaussianmarker} embeds watermarks into 3DGS through uncertainty-based perturbations for high-quality protection.
However, these NeRF and 3DGS watermarking approaches are all fine-tuning-based methods, requiring post-optimization that takes dozens of minutes or even hours.
% Although NeRF Signature~\cite{luo2025nerf} uses codebook-aided signature embedding to achieve imperceptible copyright protection without model fine-tuning, it is only applicable to the NeRF architecture.
This motivates us to develop generalizable 3DGS watermarking in a feed-forward inference manner, which takes only seconds for any 3DGS model, making it practical for real-world applications.

%% file: sec/3_preliminary.tex
\section{Preliminary}
\label{sec:preliminary}

% \paragraph{3DGS}
\noindent{\textbf{3DGS.}}
3DGS \cite{kerbl3Dgaussians} models the 3D scenes or objects as a collection of 3D Gaussians.
Each 3D Gaussian is parameterized by its mean $\mu \in \mathbb{R}^3$ as the position, opacity $\alpha \in \mathbb{R}^3$ for transparency, scaling $S$ and rotation $R$ as the shape, and spherical harmonics $c \in \mathbb{R}^{3 \times d}$ as the view-dependent color.
In our experiments, we focus on view-independent RGB color $S \in \mathbb{R}^{3}$ in the spherical harmonics for each Gaussian.
During rendering, 3DGS follows a typical neural point-based approach \cite{kopanas2022neural} to compute the color $C$ of a pixel by blending $\mathcal{N}$ depth ordered points.

% \paragraph{Splatter Images.}
\noindent{\textbf{Splatter Images.}}
While 3DGS achieves remarkable speed and visual quality, its unstructured, permutation-invariant representation poses fundamental challenges for neural network processing.
Unlike conventional grid-based data such as images, 3DGS roots in point cloud structure, which complicates its compatibility with established image-foundation neural networks \cite{he2016deep, dosovitskiy2020vit}. 
To address this, recent works \cite{tang2024lgm, xu2024grm} propose generating 3D Gaussians from multi-view images into \textbf{Splatter Images}, a structured $H \times W$ grid representation where each pixel corresponds to a 3D Gaussian parameterized by 14 channels. 
These channels encode attributes for position $\mu$, color $c$, opacity $\sigma$, rotation $r$, and scaling $s$, effectively modeling both appearance and geometry as spatially aligned image-like tensors. 
By regularizing unstructured 3D Gaussians into this grid format, Splatter Images bridge the critical gap between 2D images and 3D gaussians, enabling the application of generalizable watermarking for 3DGS.

%% file: sec/4_proposed_method.tex
\begin{figure*}[tp]
  \centering
  \begin{subfigure}{\linewidth}
    % \fbox{\rule{0pt}{40mm}\rule{170mm}{0pt}}
    \includegraphics[width=1.0\linewidth]{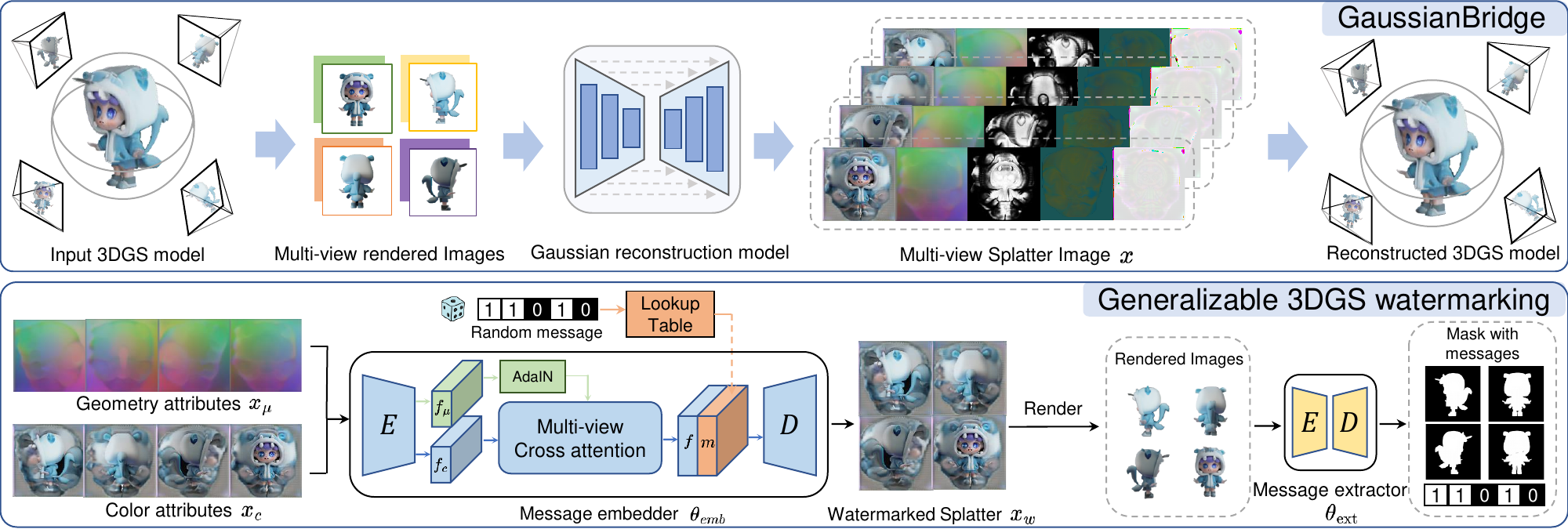}
  \end{subfigure}
  
  \caption{Overview of our proposed method. The GaussianBridge module enables bi-directional transformation between 3D Gaussian Splatting (3DGS) models and Splatter Images. We propose novel 3DGS generalizable watermarking through a two-phase approach: a message embedder that applies a selective strategy to introduce targeted perturbations in the Splatter Image, and a message extractor that generates segmentation masks to precisely locate watermarked regions and retrieve the embedded messages.}
  \label{fig:algo_overview}
\end{figure*}

\section{Proposed method}

Our generalizable watermarking for 3DGS contains two key components: 
(1) The \textbf{GaussianBridge} module to convert the unstructured 3DGS model into Splatter Images, a grid-based 3DGS representation.
(2) The \textbf{watermarking model} to embed copyright messages on the reconstructed Splatter Images and extract the embedded message from the rendered images.

\subsection{GaussianBridge}\label{sec:fine-tune}

Our whole approach is based on the Splatter Images to achieve generalizable watermarking of 3DGS. However, with many point cloud 3DGS models in existence, we still need to consider their protection. We thus introduce the GaussianBridge module with a plug-and-play property that establishes a conversion pathway between point cloud structures and grid-based Splatter Image structures. Then, the Splatter Images can be directly processed by the neural networks to achieve direct and generalizable watermarking.

As shown in \Fref{fig:algo_overview}, our GaussianBridge renders multi-view images orbitally around the 3DGS object with their camera parameters. Then, a Gaussian reconstruction model is utilized to transfer the 3DGS models from 3D point cloud formats into grid-based Splatter Images representations. 
To comprehensively preserve the information in the original 3DGS model with point-cloud structures, we propose a strategy to change the input setting of the current Gaussian reconstruction from a fixed number of images to an arbitrary number of images.

Since the GaussianBridge module leverages the Splatter Images structure to transform 3D Gaussian Splatting models, our method can be applied to per-pixel-based Gaussian reconstruction models.
Specifically, we adopt LGM \cite{tang2024lgm} as the baseline method for our Gaussian reconstruction model due to its strong performance and balance between reconstruction quality and computational efficiency.
Furthermore, our method can also be extended to more advanced Gaussian reconstruction models such as GRM and StereoGS, which enables our approach to be well-suited for both 3D Gaussian reconstruction tasks and 3D Gaussian generation tasks.

Specifically, we adapt the LGM \cite{tang2024lgm} baseline from a fixed number of input views setting to handle arbitrary input views to achieve high-quality 3DGS reconstructions.
We first densely encode camera poses using Plücker ray embedding \cite{xu2023dmv3d}, then concatenate multi-view images and ray embeddings to form a unified 9-channel feature map for 3D Gaussian prediction. 
This multi-view feature embedding $F$ is defined as: $F=\left\{C, R_o \times R_d, R_d\right\},$ where $C$ is the images color and $R_o, R_d$ represent ray origins and directions, respectively.

The asymmetric U-Net architecture in LGM takes $F$ as input and employs residual layers \cite{he2016deep} and self-attention mechanisms \cite{vaswani2017attention}, following designs from prior works \cite{ho2020denoising, metzer2023latent, szymanowicz2024splatter}. 
To enhance cross-view information propagation, we flatten and concatenate multi-view features before self-attention layers, aligning with strategies in multi-view diffusion models \cite{wang2023imagedream, shi2023mvdream}.
The final output feature map comprises $14$ channels, with each pixel representing a 3D Gaussian parameterized by position $\mu$, scaling $s$, rotation $r$, opacity $\sigma$ and color $c$. 
This structured representation aligns with the SplatterImage format \cite{szymanowicz2024splatter}, ensuring compatibility with generalizable watermarking.

\subsection{Generalizable 3DGS watermarking} \label{sec:3dgs_watermarking}

We introduce a generalizable watermarking system capable of directly watermarking 3DGS models. Unlike previous approaches that require computationally expensive fine-tuning and optimization processes, our method offers an efficient alternative for 3DGS protection.

As illustrated in \Fref{fig:algo_overview}, our proposed framework consists of a watermark embedder $\theta_{{emb}}$ and a watermark extractor $\theta_{{ext}}$.
The embedder $\theta_{emb}$ generates a watermarked Splatter Image $x_w$ by adding a perturbation $\delta$ on the original Splatter Image $x$.
The watermark extractor $\theta_{\text{ext}}$ predicts a segmentation mask $y^{mask}$ to localize watermarked regions and extracts pixel-wise embedded messages $y^{msg}$ from watermarked rendered images.

\subsubsection{Multi-view watermark embedder}

Our model accepts multi-view Splatter Images $x$ and a binary message $\hat{m}$ as input. 
Following proven image watermarking paradigms \cite{zhu2018hidden, sander2025watermark}, we adopt an encoder-decoder architecture for the embedder. 
However, 3DGS models introduce unique challenges due to their multi-attribute structure. To ensure watermark imperceptibility, we must consider both \textbf{in-attribute consistency} and \textbf{cross-attribute correlations}.
We propose a dual-strategy approach: \textit{in-attribute embedding} to encode messages within selective attributes, and \textit{cross-attribute constraint} to modulate perturbations across attributes. Such an approach leverages the inherent structure of 3DGS to achieve imperceptibility.

% \paragraph{In-attribute embedding.} 
\noindent{\textbf{{In-attribute embedding.}}}
We apply the embedder on the Splatter Image color attribute $x_c$ to generate watermarked color attributes, and keep other attributes unchanged. This selective embedding strategy ensures the efficient and effective message embedding while avoiding perturbing the 3DGS geometry \cite{luo2023copyrnerf}.
The embedder takes $x_c$ as the input and the position attribute $x_\mu$ as geometrical guidance to produce watermarked color attributes that harmonize with the underlying 3D geometry.

The embedder first encodes $x_c$ and $x_\mu$ with ResidualNet-based \cite{he2016deep} down-sampling networks into feature representations $f_c$ and $f_\mu$.
To ensure geometric consistency, we apply adaptive instance normalization (AdaIN) layers to dynamically modulate these color features $f_c$ using encoded geometry attributes $f_\mu$, enabling joint reasoning between appearance and geometry.
This modulated feature representation is further refined through a multi-view self-attention mechanism, which explicitly models inter-view dependencies and ensures the learned representation adheres to the 3DGS multi-view geometric constraints \cite{tang2024lgm, shi2023mvdream}:
\begin{equation}
    f=\operatorname{MultiViewAttn}{(\operatorname{AdaIN}(f_c, f_\mu))},
\end{equation}
where $f$ is the learned middle representation.
Message embedding occurs at the bottleneck layer, where a learnable binary lookup table maps binary messages into latent perturbations. 
These perturbations are concatenated with the features $f$, ensuring the watermark embedding process is established on both appearance and geometric priors of the 3DGS model. 
The decoder then processes the concatenated representation through additional multi-view self-attention layers and ResidualNet-based \cite{he2016deep} up-sampling networks to generate a perturbation map $\delta=\theta_{emb}(x_c, \hat{m})$.
This perturbation is applied to the original color attribute $x_c$ to produce the final watermarked color attributes $x_{cw}$, ensuring minimal perceptual distortion while embedding recoverable information:
\begin{equation}
    % S_w = \theta_{ext}(x + \theta_{emb}(x, \hat{m}))
    x_{cw} = x_c + \theta_{emb}(x_c, \hat{m}).
\end{equation}
The watermarked attribute $ x_{cw} $ replaces $ x_c $ in the original Splatter Image $ x $, producing the watermarked version $ x_w $.

% \paragraph{Cross-attribute constrains.}
\noindent{\textbf{{Cross-attribute constrains.}}}
To ensure the imperceptibility,  we propose Gaussian Uncertainty Perceptual (GUP) heatmap, 
which utilizes uncertainty to constrain the across-attributes interferences caused by the embedded watermarks.

In previous fine-tuning-based 3D watermarking \cite{li2023steganerf, jang20243dgsw}, many methods rely on reference-based methods to estimate the rendering contribution map or saliency map to constrain perturbation. 
These methods rely on the ground truth data, which is impractical in the inference situations for \textbf{3DGS generative models} \cite{tang2024lgm, xu2024grm, gslrm2024} when ground truth images are unavailable.
While uncertainty-based estimation \cite{huang2024gaussianmarker} provides a non-reference-based approach by adding perturbations into the model parameters, which is suitable for the 3DGS generative models without referencing ground truth images.
Thus, we use the uncertainty estimation methods \cite{Jiang2024FisherRF, huang2024gaussianmarker} for constraining watermark perturbation.
We use the simplification of the Hessian matrix $\mathbf{H}$ as the approximated Fisher information to estimate uncertainty values $\mathbf{U}$ for each Gaussian \cite{kirsch2022unifying, huang2024gaussianmarker}:
$
\mathbf{H} = \nabla_{\mathcal{G}}I_{\mathcal{G}} \nabla_{\mathcal{G}}I_{\mathcal{G}}^T,
$
where $\nabla_{\mathcal{G}}$ is the gradient for the parameters of 3DGS model $\mathcal{G}$ and $I$ is the rendered image.
We estimate $\mathbf{U}$ by computing logarithm of $\mathbf{H}$ for rearrange the exponential large values:
\begin{equation}
\mathbf{U} = \log (\nabla_{\mathcal{G}}I_{\mathcal{G}} \nabla_{\mathcal{G}} I_{\mathcal{G}}^T).
\label{eq:uncertainty}
\end{equation}
Specifically, we exclude rotations $r$ from $\mathcal{G}$ because the changes in this attribute can easily cause noticeable differences such as needling artifacts on the boundary regions.

Given the uncertainty values $\mathbf{U}$, we can generate GUP heatmaps: $\gamma = \mathcal{R}(\mathbf{U}, v)$, where $\mathcal{R}$ is the splatting rendering function \cite{Jiang2024FisherRF} and $v$ is a normalized viewing camera parameters for Splatter Images.
These GUP heatmaps $\gamma$ are modulated on Splatter Images to constrain perturbations, thus ensuring imperceptible watermarking:
\begin{equation}
x_{w}=x+\alpha \cdot \gamma \odot \delta,
\label{eq:gup_heatmap}
\end{equation}
where $\gamma$ is the GUP heatmaps to modulate the perturbation $\delta$, and $\alpha$ is the scaling factor to adapt $\delta$.

\subsubsection{Robust watermark extractor}

We design our watermark extractor as a segmentation-based approach that simultaneously localizes watermarked regions and retrieves embedded messages. 
Following the watermark-anything model (WAM) \cite{sander2025watermark}, our architecture combines a Vision Transformer (ViT) \cite{dosovitskiy2020vit} encoder for global feature extraction and a pixel decoder that upsamples features to match the input resolution, producing per-pixel outputs $y = \{y^{mask}, y^{msg}\}$ of size $1+N$ bits. This design eliminates the need for an auxiliary discriminator \cite{li2023steganerf, huang2024geometrysticker}.

Previous methods \cite{huang2024geometrysticker, jang2024waterf, jang20243dgsw, huang2024gaussianmarker} often rely on HiDDeN’s extractor \cite{zhu2018hidden}, which struggle to accurately retrieve watermarks when applied to 3D models, as watermarked pixels frequently occupy sparse regions under varying camera viewpoints.
Our watermark extractor $\theta_{ext}$ localizes and extracts embedded messages from watermarked rendered images $I_w$ using:
\begin{equation}
I_w = \mathcal{R}(x_w, V);~~~~~~y^{mask}, y^{msg} = \theta_{ext}(I_w)
\label{eq:msg_extraction}
\end{equation}
where $\mathcal{R}$ represents the rendering function, $V$ denotes the viewing camera, $y^{mask}$ is the segmentation mask that localizes watermarked regions, and $y^{msg}$ represents the pixel-wise embedded messages.

Furthermore, most 3D watermarking methods \cite{luo2023copyrnerf, jang2024waterf} cannot verify copyright messages across both model parameters and rendered images. 
Our method extracts watermarks from 3DGS models via Splatter Images without scene-specific tuning. 
Specifically, the watermark embedded in the watermarked Splatter Image color attribute $x_c$ can be directly extracted by our message extractor $\theta_{ext}$, thus ensuring the generalizability for embedding and extracting from 3DGS model parameters.

\noindent{\textbf{Augmentation}.}
Our augmentation pipeline simulates real-world distortions on 2D rendered images via geometric transformations (resize, scaling, rotation, translation, perspective, crop-out) and photometric adjustments (noise, JPEG compression, blur, brightness/contrast shifts), reflecting typical degradations on real-world applications to obtain the watermarking robustness.
To ensure 3D robustness, we also introduce 3D-aware augmentations (geometric edits: cropping/translation/rotation; noise injection) on Splatter Images, preserving extraction accuracy even when malicious alter the 3D representations.

% \paragraph{\textbf{Training losses.}} 
\subsection{Training losses}
The training minimizes the objective function $\mathcal{L}_{total}$ which is a linear combination of the mask and message losses: 
$\mathcal{L}_{total} = \lambda_1 \cdot \mathcal{L}_{mask} + \lambda_2 \cdot \mathcal{L}_{msg}$.
The mask loss is the average of the pixel-wise cross-entropy between detecting mask $y^{mask}$ and the ground truth $\hat{y}^{mask}$ to predict whether a pixel contains watermark message or not. 
\begin{equation}
\mathcal{L}_{mask}=-(y^{mask} \log (\hat{y}^{mask})+(1-y^{mask}) \log (1-\hat{y}^{mask})) .
\end{equation}
Similarly, the message loss is the average of the pixel-wise and bit-wise binary cross-entropy between the predicted watermark messages $y^{msg}$ and the ground truth watermark messages $\hat{m}$:
\begin{equation}
\mathcal{L}_{msg}=-(y^{msg} \log (\hat{m})+(1-y^{msg}) \log (1-\hat{m})) .
\end{equation}

% The detection loss is the average of the pixel-wise cross-entropy between $\ydet(\theta)\in[0,1]^{h \times w}$ and the ground truth $\ydetgt\in\{0,1\}^{h \times w}$ (pixel watermarked or not).
% Similarly, the decoding loss is the average of the pixel-wise and bit-wise binary cross-entropy between $\ydec_{i,k}(\theta)$ and $m_k$ only over the watermarked pixels, where $m$ is a random message originally embedded in that image.
% For a given image and message, $\ell_\detection$ and $\ell_\decoding$ are:
% \begin{align}
%     \ell_\detection(\theta) &= 
%     \frac{-1}{h \times w} \sum_{i=1}^{h\times w} \left[ \ydetgt_i \log(\ydet_i(\theta)) + (1 - \ydetgt_i) \log(1 - \ydet_i(\theta)) \right], \\
%     \ell_\decoding(\theta) &= 
%     \frac{-1}{\nbits \times \sum_{i=1}^{h\times w} \ydetgt_i} \sum_{i=1}^{h\times w} \ydetgt_i \sum_{k=1}^{\nbits} \left[ m_k \log(\ydec_{i,k}(\theta)) + (1 - m_k) \log(1 - \ydec_{i,k}(\theta)) \right].
% \end{align}

% \paragraph{\textbf{Training Losses}.}

%% file: sec/5_experiments.tex
\begin{figure*}[htbp]
  \centering
  \begin{subfigure}{\linewidth}
    % \fbox{\rule{0pt}{40mm}\rule{170mm}{0pt}}
    \includegraphics[width=1.0\linewidth]{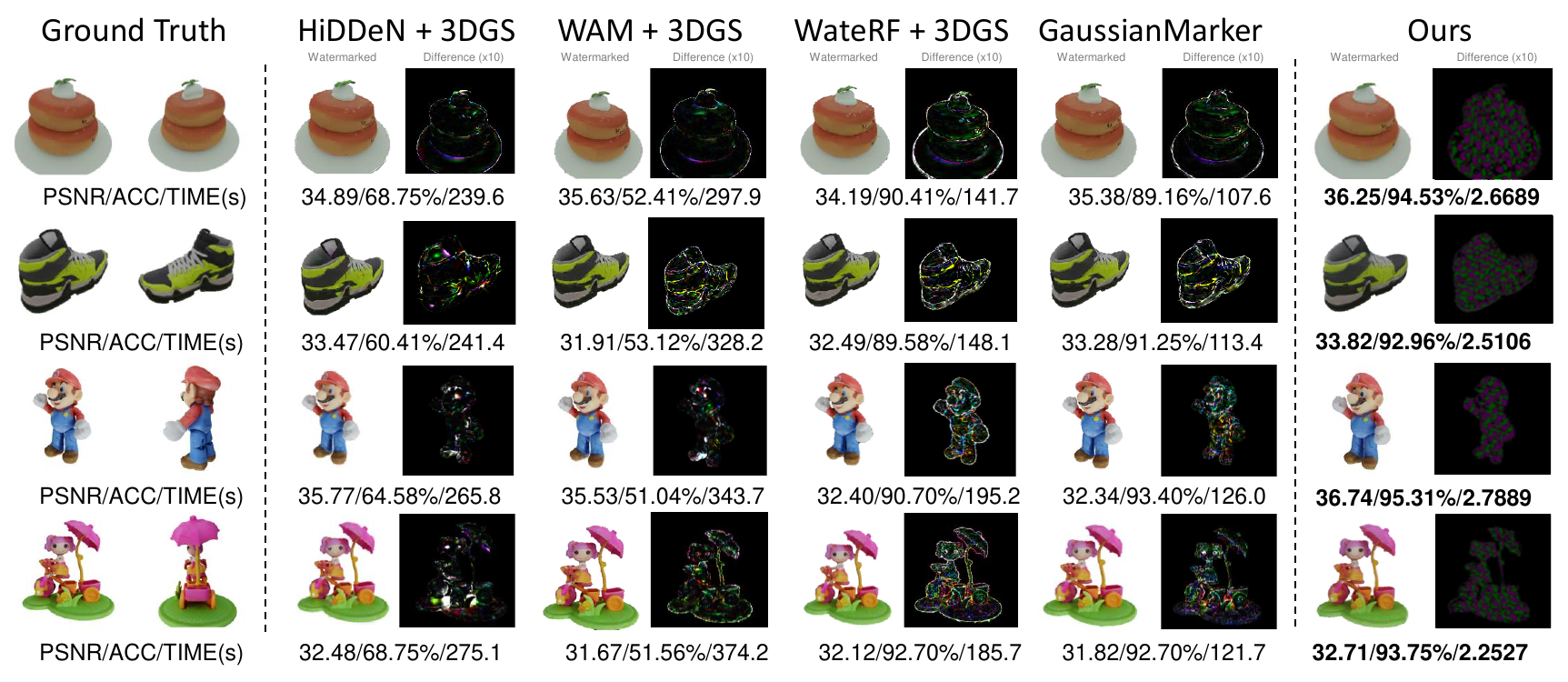}
  \end{subfigure}
  \vspace{-20pt}
  \caption{Qualitative results of comparing our method with the baselines on the evaluation dataset. The extraction accuracies are reported in 32 bits. The differences between watermarked and ground truth images are amplified by 10 times for better visualization.}
  \label{fig:qualitative-obj-gso}
  % \label{fig:sparse_view_results}
\end{figure*}

\section{Experiments}
\label{sec:experiments}

\subsection{Implementation details}
\label{exp:implementation_details}

% \paragraph{Dataset.}
% \subsubsection{Dataset.}
\noindent\textbf{Dataset.}
We choose the Objaverse dataset \cite{deitke2023objaverse} as our training dataset to train our watermarking models.
Specifically, we utilize the LVIS annotation to obtain a subset of Objaverse \cite{deitke2023objaverse}, which contains around $45$k high-quality objects.
For each object in the dataset, we render $512 \times 512$ images from 32 random viewpoints. 
The 32 rendered views from different angles are then used to train 3DGS models for 10K iterations using \cite{kerbl3Dgaussians}.
This way, we create a high-quality and large-scale 3DGS dataset of $45K$ objects and 144k rendering views from Objaverse \cite{deitke2023objaverse}, which is sufficient to train our GaussianBridge module and watermarking models.
% To be updated
We evaluate the qualitative and quantitative results for reconstruction and watermarking performance using Objaverse \cite{deitke2023objaverse} and Google Scanned Objects (GSO) \cite{downs2022google} datasets.
For the Objaverse dataset, we randomly sample 200 objects from the LVIS annotation as evaluation dataset and others as training dataset.
The GSO dataset comprises around 1,000 objects, from which we randomly select 50 objects as our evaluation dataset. 
Specifically, we render 16 images of each object in the GSO evaluation set in an orbiting trajectory with uniform azimuths varying positive elevations in $\{0^{\circ}, 20^{\circ}\}$ for sampling on the top semi-sphere of an object.

% For the 3DGS generation model, we use pre-trained weights of LGM \cite{lgm} without fine-tuning or modification.

% After fitting all the object to 3DGS representation, we convert the objects to the corresponding UV maps (\ie, UVGS) through Spherical Mapping as illustrated in Fig.~\ref{fig:architecture}. 
% For the course of our experiments, we only mapped the objects to a single layer UV maps as it was sufficient to represent the general purpose Objaverse objects with minimal quality loss. 
% Through mapping, we gathered a large UVGS dataset of  $\mytilde 400K$ maps.
% We fix the size of the UV maps to $512\times512$. Through our experiments, we found that UV maps of size $512 \times 512$ are sufficient to represent objects in our dataset and capable of storing upto $262K$ unique Gaussians.

% \paragraph{Training settings.} 
% \subsubsection{Training settings.}
% Similar to LRM \cite{hong2023lrm}, we transform the cameras of each batch such that the first input view is always the front view with an identity rotation matrix and fixed translation.
% In specific, we set the size of reconstructed Splatter Image to $128\times128$, which   
% For all baselines, we use four input views with camera elevation angle = 0 and azimuths degrees = $\{0,90,180,270\}$ to cover the entire object and evaluate the reconstruction quality on the remaining 12 views.
\noindent\textbf{Training settings.}
Our training contains two stages, and all experiments are trained on $8$ NVIDIA V100 (32G) GPUs for about 1 day.
In the first stage, we train our GaussianBridge module to obtain the high-quality multi-view Splatter Image representations.
We use LGM \cite{tang2024lgm} as the baseline for 3D Gaussian reconstruction model and use 4 images as the default number of views and enabling to increase in the number of views up to 8. 
The size of the reconstructed Splatter Image is $128 \times 128$ for each input view, and the 3D Gaussians are rendered at $512 \times 512$ resolution.

In the second stage, we utilize the GaussianBridge module for obtaining multi-view Splatter Images from the 3DGS models in the training dataset to train our watermarking models.
For any 3DGS model, we render 4 images with camera elevation angle = $0$ and azimuths degrees = $\{0,90,180,270\}$ to cover the entire object. 
The GaussianBridge module uses the rendered images to reconstruct multi-view Splatter Images.
We apply our multi-view watermark embedder on the Splatter Images and then randomly select 8 camera views to render watermarked novel views for message extraction.
We set the training batch size = 4 and use AdamW optimizer for the multi-view watermark embedder. 
The training epoch is set to 30 epochs. 
The learning rate is linearly increased from $10^{-6}$ to $10^{-5}$ over the first 5 epochs, and then follows a cosine schedule down to $10^{-7}$ until epoch 30. 
% We initialize the our message extractor with the pre-trained weight in \cite{sander2025watermark}  
% Since the rendered images from the Objaverse 3DGS models slightly differ from the SA-1B dataset \cite{kirillov2023segment} trained for WAM \cite{sander2025watermark}, we set a constant small learning rate at $10^{-7}$ to fine-tune the message extractor.
We apply the augmentation pipeline to ensure robustness during training.
We also extract messages from the watermarked Splatter Images to ensure robustness in both 2D rendered images and 3D Gaussians parameters.

\begin{table*}[htbp]
\centering
\caption{
Reconstruction qualities and bit accuracy compared with different baselines.
PSNR, SSIM and LPIPS are computed between the watermarked images and ground truth images.
% The Gaussian noise's standard deviation is set to $0.1$; the rotation degrees are set between $\pm\pi/6$; the scaling sets a random scale factor below $25\%$; and the Gaussian blur standard deviation is set to $0.1$. 
The results are computed on the average of samples from Objaverse and GSO datasets.
}
% \vspace{-10pt}
\label{table:watermark_images_different_bits}
\begin{adjustbox}{width=0.9\textwidth}
\begin{tabular}{c|c|ccccc|ccccc}
\toprule
\multirow{2}{*}{Dataset} & \multirow{2}{*}{Method} & \multicolumn{5}{c|}{32 bit} & \multicolumn{5}{c}{48 bit} \\
& & \multirow{1}{*}{Bit Acc$\uparrow$} & \multirow{1}{*}{PSNR$\uparrow$} & \multirow{1}{*}{SSIM$\uparrow$} & \multirow{1}{*}{LPIPS$\downarrow$} & \multirow{1}{*}{Time(s)$\downarrow$} & \multirow{1}{*}{Bit Acc$\uparrow$} & \multirow{1}{*}{PSNR$\uparrow$} & \multirow{1}{*}{SSIM$\uparrow$} & \multirow{1}{*}{LPIPS$\downarrow$} & \multirow{1}{*}{Time(s)$\downarrow$} \\
\midrule
\multirow{5}{*}{Objaverse + GSO} & HiDDeN $+$ 3DGS & 66.03 & 33.71 & 0.9022 & 0.0782 & 255.34 & 63.19 & 32.87 & 0.9041 & 0.0995 & 269.86 \\ 
                           & WAM $+$ 3DGS & 52.03 & 33.95 & 0.8929 & 0.0801 & 336.61 & 50.84 & 32.82 & 0.8896 & 0.1005 & 347.53 \\ 
                           & WateRF $+$ 3DGS & 90.84 & 32.63 & 0.8973 & 0.0951 & 167.61 & 90.14 & 31.40 & 0.8716 & 0.0939 & 175.35 \\ 
                           & GaussianMarker & 91.62 & 33.20 & 0.9082 & 0.0822 & 117.14 & 90.93 & 32.69 & 0.8909 & 0.0965 & 126.08 \\ 
                           % & WAM $+$ Splatter & 28.17/0.9047 & 0.0878  & xxx & 67.13 \\ 
                           & Ours  & \textbf{94.41} & \textbf{35.12} & \textbf{0.9234} & \textbf{0.0775} & \textbf{2.53} & \textbf{93.36} & \textbf{33.99} & \textbf{0.9189} & \textbf{0.0891} & \textbf{2.65} \\ 
% \midrule
% \multirow{5}{*}{Google Scan Object} & HiDDeN $+$ 3DGS & 22.47/0.8053 & 0.4825 & xxx & xxx/xxx & xxx & xxx \\       
%                                     & WAM $+$ 3DGS & 27.20/0.8151 & 0.2143 & xxx & xxx/xxx & xxx & xxx \\  
%                                     & WateRF $+$ 3DGS & 24.84/0.7992 & 0.1705 & xxx & xxx/xxx & xxx & xxx \\ 
%                                     & GaussianMarker  & 27.04/0.8452 & 0.1357 & xxx & xxx/xxx & xxx & xxx \\ 
%                                     % & WAM $+$ Splatter & 28.17/0.9047 & 0.0878  & xxx & 67.13 & 67.06 & 63.43 & 64.04 & 66.38 \\
%                                     & Ours & \textbf{29.16}/\textbf{0.8808} & \textbf{0.1197} & \textbf{xxx} & xxx/xxx & xxx & xxx \\ 
\bottomrule
\end{tabular}
\end{adjustbox}
% \vspace{-5mm}
\end{table*}

\begin{table*}[htbp]
\centering
\caption{
Quantitative assessment of robustness against various attacks, compared to baseline methods. 
The reported results are averaged across the Objaverse and GSO datasets. All experiments were conducted using 32-bit messages.
% Reconstruction qualities and bit accuracy on 32 bits compared with different baselines.
% PSNR/SSIM and LPIPS are computed between the original and watermarked rendered images.
% The Gaussian noise's standard deviation is set to $0.1$; the rotation degrees are set between $\pm\pi/6$; the scaling sets a random scale factor below $25\%$; and the Gaussian blur standard deviation is set to $0.1$. 
The results are computed on the average of all examples.
}
% \vspace{-20pt}
\label{table:watermark_images_different_distortions}
\begin{adjustbox}{width=0.95\textwidth}
\begin{tabular}{c|c|ccccccccccc}
\toprule
\multirow{3}{*}{Dataset} & \multirow{3}{*}{Method} & \multicolumn{8}{c}{Bit accuracy $\uparrow$ (\%)} \\
&  & None & Noise & JPEG & Scaling & Blur & Rotation & Translation & Crop-out \\
&  &      &  $(\nu=0.1)$   & $(Q = 40)$ & $(sc\leq 70 \%)$ & $(\xi=0.1)$ & $(ro \leq \pm \pi / 6)$ & $(t\leq20\%)$ & $(cr \leq 50\%)$ \\
\midrule
\multirow{4}{*}{Objaverse + GSO } & HiDDeN $+$ 3DGS & 66.03 & 64.71 & 53.34 & 50.86 & 65.24 & 59.15 & 64.83 & 52.17 \\ 
                         & WateRF $+$ 3DGS & 90.84 & 86.23 & 75.80 & 58.08 & 89.64 & 85.42 & 89.43 & 57.67\\ 
                         & GaussianMarker  & 91.62 & 90.98 & 71.66 & 61.50 & 87.75 & 81.32 & 87.75 & 59.15 \\ 
                         & Ours   &  \textbf{94.35} & \textbf{93.20} & \textbf{89.06} & \textbf{89.95} & \textbf{92.31} & \textbf{92.45} & \textbf{91.88} & \textbf{87.08} \\ 
\bottomrule
\end{tabular}
\end{adjustbox}
% \vspace{-5mm}
\end{table*}

\subsection{Experimental settings}
\label{exp:Experiment settings}

\noindent\textbf{Baselines.}
We design experiments to validate the message extraction on both rendered 2D images and 3D Gaussian parameters, demonstrating the effectiveness of our proposed method.
For message extraction from rendered images, we compare our proposed method with four baselines for a fair comparison:
1) \textbf{Fine-tuning 3DGS+HiDDeN} \cite{zhu2018hidden}:  Preprocessing images with the classical image watermarking method HiDDeN \cite{zhu2018hidden} before the training of 3DGS \cite{kerbl3Dgaussians};
2) \textbf{Fine-tuning 3DGS+WAM} \cite{zhu2018hidden}:  Preprocessing images with the state-of-the-art image watermarking method HiDDeN \cite{sander2025watermark} before the training of 3DGS \cite{kerbl3Dgaussians}; 
3) \textbf{3DGS+WateRF}: 
Train a watermarked 3DGS object using frequency-based watermarking method via WateRF \cite{jang2024waterf} approach;
4) \textbf{GausssianMarker}: 
state-of-the-art fine-tuning-based 3DGS watermarking method via uncertainty estimation;
For 3D message extraction, we compare our method with the GaussianMarker baseline since other methods do not provide a 3D watermarking method.
% 5) \textbf{Splatter+WAM}:
% We use our Gaussian reconstruction module to obtain multi-view Splatter Images and directly apply the WAM \cite{sander2025watermark} model on the color attribute for watermarking.
To illustrate our method can be generalizable to different 3D Gaussian generative models, we select three baselines, including DreamGaussian \cite{tang2023dreamgaussian}, Triplane Gaussians \cite{zou2024triplane} and LGM \cite{tang2024lgm}.
To ensure a fair comparison, we first build the point cloud 3DGS model similar to the baselines. Then, we use our pre-trained GaussianBridge to convert 3DGS from the point cloud to Splatter Images.
%\renjie{To ensure a fair comparison, we first train the point cloud 3DGS model via \cite{kerbl3Dgaussians}. Then, we use our trained GaussianBridge to convert 3DGS from the point cloud to Splatter Images. }
%TODO: add trellis

\noindent\textbf{Evaluation methodology.}
We set the bit length of binary messages to $32$ and $48$ bits to test \textit{capacity} for all baselines and our method.
We evaluate the performance of our proposed method by comparing it with other digital watermarking baselines using the standard of imperceptibility and robustness. 
% For \textit{capacity}, we set the bit length of copyright messages to $32$ bits, aligning with the maximum length previously employed in 3D model watermarking methods~\cite{yoo2022deep, luo2023copyrnerf}. 
For \textit{imperceptibility}, we evaluate the reconstruction quality with PSNR, SSIM, and LPIPS \cite{Zhang_Isola_Efros_Shechtman_Wang_2018} for rendered images. 
For \textit{robustness}, we evaluate whether the binary messages in rendered images can remain consistent against various distortions, 
including 2D Gaussian noise with standard deviation $\nu=0.1$, 
JPEG compression with quality factor $Q=40$, 
scaling with the size factor $sc\leq70\%$, 
and Gaussian blur with kernel size $k=3$ and the standard deviation $\xi=0.1$. 
We also evaluate whether the copyright messages in 3D Gaussians can remain consistent against various 3D attacks, including 3D Gaussian noise with standard deviation $\nu'=0.1$, translation with the ratio factor $t \leq 20\%$, rotation with the angular factor $ro \leq \pm \pi / 6$, and crop-out with the cropping size percentage $cr \leq 50\%$.

\subsection{Experimental results}

\noindent\textbf{Qualitative results.} 
We use PSNR, SSIM, and LPIPS \cite{Zhang_Isola_Efros_Shechtman_Wang_2018} to measure the Gaussian reconstruction quality.
\Fref{fig:qualitative-obj-gso} demonstrates the qualitative and quantitative reconstruction results on the Objaverse and GSO datasets. 
We compare the reconstruction qualities and bit accuracies with all baselines, and the qualitative results are shown in \Fref{fig:qualitative-obj-gso}.
Both \textbf{HiDDeN $+$ 3DGS} and \textbf{WAM $+$ 3DGS} watermark the training images but struggle to decode the watermark messages, which align with previous methods \cite{luo2023copyrnerf, huang2024gaussianmarker} since it is difficult to transmit the watermark signal from the 2D images into the 3DGS models. 
Both \textbf{WateRF $+$ 3DGS} and \textbf{GaussianMarker} are all fine-tuning-based methods. Although these methods show promising decoding accuracy, they require per-object optimization for more than 100 seconds and also exhibit vulnerability when subjected to zoomed-in/out operations, making them impractical for real-world applications.

\noindent\textbf{Quantitative results.}
We display the quantitative results in \Tref{table:watermark_images_different_bits} and \Tref{table:watermark_images_different_distortions}.
Our method shows superior performance on both reconstruction quality and decoding accuracy since we consider the multi-view corresponding and the 3D Gaussian geometry condition.
Even with different distortions to the rendered images, our method can still achieve high decoding accuracy to reliably safeguard the 3DGS models.
MarkSplatter can also extract the message from the 3D Gaussian attribute via the format of Splatter Image.
We compare our method with the GaussianMarker\cite{huang2024gaussianmarker} method in \Tref{table:3d_watermark}. 
GaussianMarker \cite{huang2024gaussianmarker} relies on a per-scene specific decoder for the 3D watermark decoding, which is impractical for real-world applications. 
Our MarkSplatter can extract watermark in 3D Gaussians in a generalizable method, which shows superior performance than the fine-tuning methods.

\begin{figure}[tp]
  \centering
  \begin{subfigure}{\linewidth}
    % \fbox{\rule{0pt}{40mm}\rule{80mm}{0pt}}
    \includegraphics[width=1.0\linewidth]{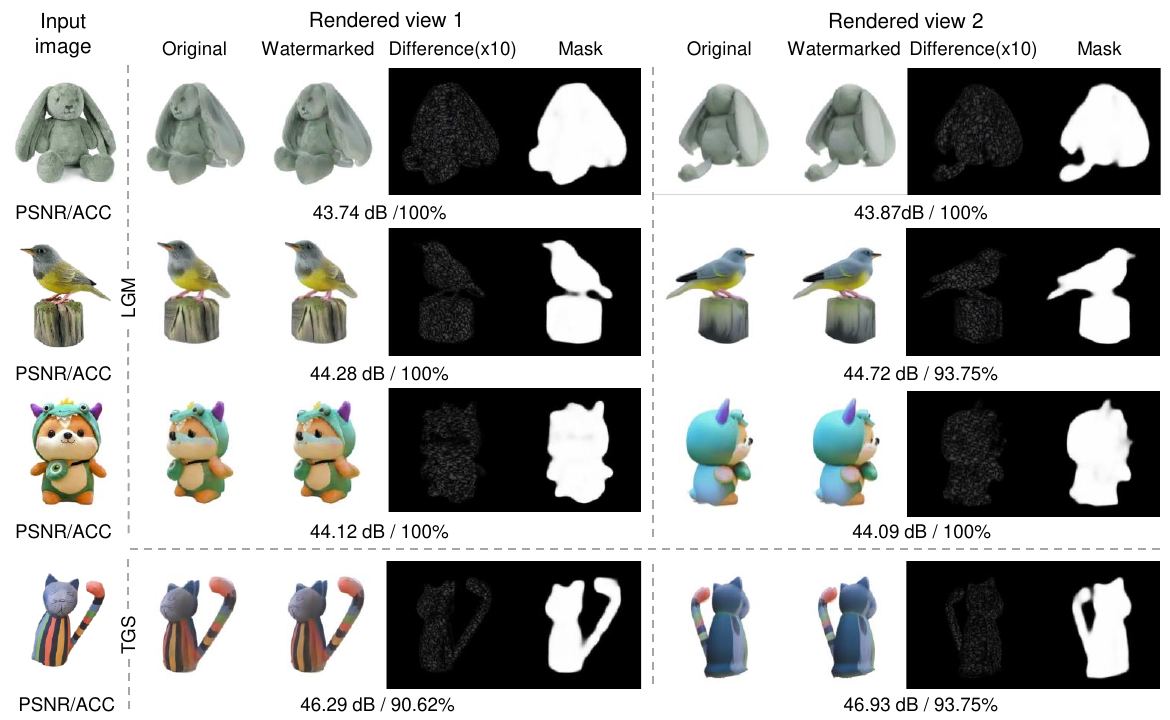}
  \end{subfigure}
  
  \caption{Watermarking performance on generative models.
  PSNR is computed between the original and watermarked rendered images. 
  The results are averaged across the given scenes and bit accuracy is in 32 bits.}
  \label{fig:qualitative_generative}
  % \label{fig:sparse_view_results}
\end{figure}

\noindent\textbf{Single-Image-to-3D results.} 
We evaluate our MarkSplatter with the Single-Image-to-3D generative models \cite{zou2024triplane, tang2024lgm} in \Fref{fig:qualitative_generative}. 
In the first column, we display the input single image. 
In the second and third columns, we show the original generated image, the watermarked image, the difference, and the predicted mask for two different viewing angles.
Specifically, in the first 3 rows, we use the LGM~\cite{tang2024lgm} pipeline to achieve Single-Image-to-3D, and \method~can directly apply to the LGM~\cite{tang2024lgm} model since its Splater Image Structure.
In the fourth row, we use the TriplaneGaussian to reconstruct a point cloud-based 3DGS model and use our GaussianBridge to convert it into a Splatter Image-based 3DGS model for \method~watermarking protection.
These results showcase the generalizability of MarkSplatter to protect any 3DGS models in point-cloud format and Splatter Image format.
% Our MarkSplatter utilizes GaussianBridge module to obtain Splatter Images for applying generalizable 3DGS watermarking.
% Furthermore, even with various 3DGS generative models, our method can still robustly decode the watermark messages from various generative models, showing that our capability of protecting the copyright of 3D Gaussian generative models in a plug-and-play method without influence its inner structure.

% \begin{figure}[tp]
%   \centering
%   \begin{subfigure}{\linewidth}
%     \fbox{\rule{0pt}{40mm}\rule{80mm}{0pt}}
%     % \includegraphics[width=1.0\linewidth]{assets/generative_results_0.3.pdf}
%   \end{subfigure}
  
%   \caption{Gaussian uncertainty perceptual heatmap.}
%   \label{fig:qualitative_generative}
%   % \label{fig:sparse_view_results}
% \end{figure}

\begin{table}[htbp]
\centering
\caption{Message extraction from 3DGS model parameters.
The results are in 32 bits and computed on the average of samples from Objaverse \cite{deitke2023objaverse} and GSO \cite{downs2022google} datasets.
}
\label{table:3d_watermark}
\begin{adjustbox}{width=0.45\textwidth}
\begin{tabular}{c|cccc|c}
\toprule
Method          & None & Noise & Rotation & Crop-out & Time(s) \\
\midrule 
GaussianMarker  &  \textbf{100\%}  & 96.48\%  & 95.44\% & 58.28\% & 680.64\\ 
Ours            &  \textbf{100\%}  & \textbf{98.66\%}  & \textbf{97.70\%} & \textbf{93.23\%} & \textbf{2.35}  \\
\bottomrule
\end{tabular}
\end{adjustbox}
% \vspace{-5mm}
\end{table}

% \vspace{-10p}
\subsection{Ablation study}

% \paragraph{Model components.}

% We ablate different components of our full model in \Tref{}.
% We first ablate the cross-attention module and apply the watermark embedder on the Splatter Images individually.
% We then ablate the ADALN module for position maps for the position-aware 3D watermarking.
% Experiment results show that the full model can achieve the most comprehensive results.
\noindent{\textbf{Model components.}}
We conduct a series of ablation studies to evaluate the contributions of individual components in our full model, as detailed in \Tref{tab:ablation_model_component}. 
% We ablate our proposed 3DGS JND model which achieves imperceptibility. 
% Without the JND heatmap, the bit accuracy shows a bit of improvement, which is due the visibility of embedded watermark is not constrained, thus making the message extraction more accurate.
% However, this sacrifices the watermarked image quality, making the watermark visible thus undermining the imperceptibility.
% Applying our 3DGS JND heatmap can constrain the watermark embedded on the Splatter Images while still maintaining high accuracy for watermark decoding.
% We remove the cross-attention module which learns the correspondence relationship between multi-view feature embeddings. 
% We ablate the AdaIN module, which is responsible for generating position maps, which are critical for enabling position-aware 3D watermarking. 
% Our experimental results demonstrate that the full model, incorporating all components, achieves the most comprehensive and robust performance compared to the ablated variants.
% We evaluate our GUP heatmap for Splatter Image which achieves imperceptibility. 
% When the GUP heatmap was removed, bit accuracy marginally improved, which is an expected outcome since unrestricted watermark embedding facilitates easier message extraction. 
% However, this improvement comes with the degradation of image quality, which makes watermarks visible and compromises the requirement of imperceptibility. 
% Our GUP heatmap effectively addresses this trade-off by strategically limiting watermark embedding on Splatter Images while preserving high bit accuracy.
Our GUP heatmap ensures imperceptible watermarks in Splatter Images. Removing the heatmap slightly boosts bit accuracy (as unrestricted embedding simplifies message extraction), but severely degrades image quality by making watermarks visible. 
The GUP heatmap balances this trade-off for watermark embedding to preserve visual quality while maintaining high decoding accuracy.
We ablate the cross-attention module and the AdaIN module and experimental results shows degraded performance, confirming their essential roles in guaranteeing message extraction accuracy by integrating spatial information across different viewpoints.
Our experiments demonstrate that the full model, incorporating all components, achieves the most comprehensive and robust performance compared to the ablated variants.

\begin{table}[h]

\centering
% \scriptsize
\caption{Ablation study for model components.
The results are in 32 bits and computed on the average of samples from Objaverse \cite{deitke2023objaverse} and GSO \cite{downs2022google} datasets.
}
% We evaluate the contribution of different MASt3R components by systematically removing each corresponding feature during the training process.
\label{tab:ablation_model_component}
\begin{adjustbox}{width=0.45\textwidth}
\begin{tabular}{l|cccccc}
\toprule
Setting & PSNR\textuparrow & SSIM\textuparrow & LPIPS \textdownarrow & Bit Acc\textuparrow \\
% \hline
\midrule
full         & \textbf{35.12} & \textbf{0.9234} & \textbf{0.0775} & 94.41\% \\
w/o GUP                 & 34.09 & 0.9132 & 0.1053 & \textbf{95.12\%} \\
w/o cross-attention     & 32.42 & 0.8961 & 0.0939 & 85.25\% \\
w/o AdaIN               & 33.30 & 0.9044 & 0.0914 & 82.12\% \\
% w/o $D'$ and $F$          & 22.79 & 0.9069 & 0.1314 & 0.1261 \\
% w/ projected depth $D$    & 21.92 & 0.8555 & 0.1200 & 0.1227 \\
\bottomrule
\end{tabular}
\end{adjustbox}
\end{table}

\noindent{\textbf{Number of Splatter Image.}}
To enhance watermark robustness, we investigate increasing the number of input Splatter Images for watermark embedding. 
While our baseline uses 4 views, we increase the rendered images for GaussianBridge module to generate 6 Splatter Images, which can improve novel-view rendering while maintaining bit accuracy.
To train our model for 6 input views, we fine-tune our GaussianBridge module previously trained with 4 input view images.
During training, we still randomly select 8 camera views for each object and set the first 6 images as the input images.
Since the larger number of input view images, each GPU can use a batch size of 4, resulting in a total batch size of 32. 
The output feature map is still at the size of $128 \times 128$ for each input view, resulting in output 3D Gaussians to $128 \times 128 \times 6 = 98304$ in total.
We use the same learning rate setting for training 4 input views, and the training can be finished in 20 epochs. 

We discuss both qualitative and quantitative results when increasing the input number of views from 4 images to 6 images. 
For the 4-input-views condition, we maintain the same camera pose setting with elevations of 0 and azimuths at $[0, 90, 180, 270]$ degrees. 
For the 6-input-views condition, we set the camera pose with elevations of 0 and azimuths of $[0, 60, 120, 180, 240, 300]$ degrees. 
The reconstruction quality is still evaluated at azimuths of $[45, 135, 225, 315]$.

\begin{figure}[htbp]
  \centering
  \begin{subfigure}{\linewidth}
    % \fbox{\rule{0pt}{40mm}\rule{170mm}{0pt}}
    \includegraphics[width=1.0\linewidth]{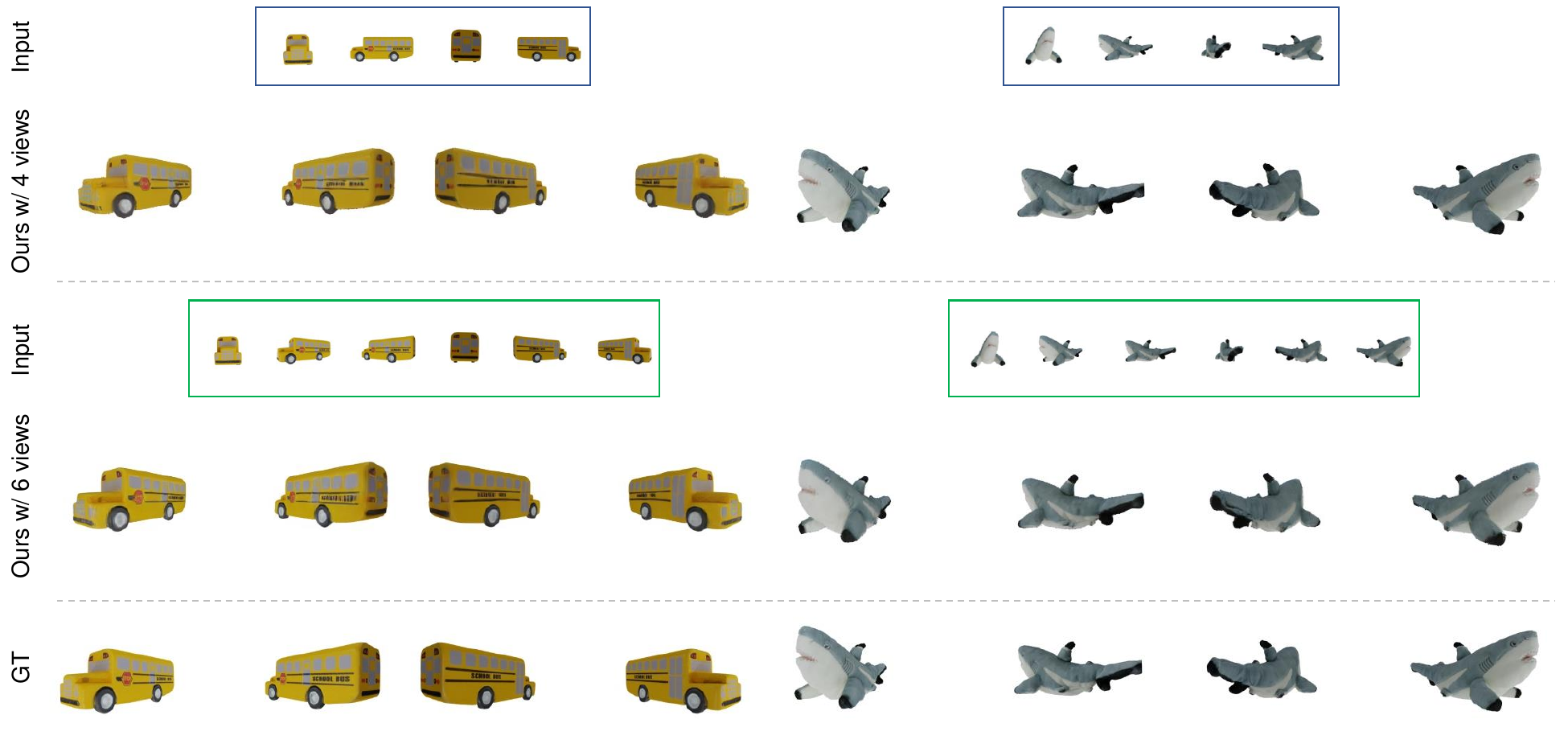}
  \end{subfigure}
  
  \caption{\textbf{GaussianBridge reconstruction with 4 input views and 6 input views.}}
  \label{fig:sparse_view_results_supp}
\end{figure}

We demonstrate qualitative results in \Fref{fig:sparse_view_results_supp}. 
With accurate camera poses and consistent input images, both our models for 4-input-views and 6-input-views can faithfully reconstruct the 3D objects.
Moreover, by leveraging six input views with denser azimuth coverage, our model achieves enhanced 3D reconstruction quality. 
As evidenced by the quantitative results in Table~\ref{tab:sparse_view_appearance_supp}, this multi-view configuration yields a marked improvement in rendering quality metrics and an enhancement in bit recovery accuracy. 
These findings demonstrate that increasing observational viewpoints can strengthen the geometric consistency of 3DGS reconstructions to improve watermark extraction robustness.

\begin{table}[h]
\centering
% \scriptsize
\caption{Evaluation for Gaussian reconstruction and watermarking quality on Objaverse and GSO datasets~\cite{downs2022google}.}
\label{tab:sparse_view_appearance_supp}

\begin{adjustbox}{width=0.4\textwidth}
\begin{tabular}{l|cccccc}
\toprule
Methods & PSNR\textuparrow & SSIM\textuparrow & LPIPS\textdownarrow  & Acc\textuparrow \\
\midrule
Ours w/ 4 views &  35.12 & 0.9234 & 0.0775 & 94.41 \\
Ours w/ 6 views &  36.20 & 0.9331 & 0.0741 & 94.80 \\
\bottomrule
\end{tabular}
\end{adjustbox}
\end{table}

%% file: sec/6_conclusion.tex
\section{Conclusion}

We propose \method, the first generalizable 3DGS watermarking framework for 3DGS models through our GaussianBridge module to encode 3DGS models into compact Splatter Images with high reconstruction quality.
% We design a multi-view watermark embedder to embed messages on the input Splatter Images and constrain the message perturbation by GUP heatmap for imperceptibility.
% A robust message extractor can localize and extract embedded messages on the watermarked regions.
A multi-view watermark embedder hides messages in Splatter Images, using a GUP heatmap to limit visible changes, and a robust extractor then locates and reads the messages.
This pipeline establishes a foundation for generalizable 3DGS watermarking without compromising reconstruction quality.

% \paragraph{Limitations and future works.}
% Our method focuses on protecting general 3DGS models, which are in a non-topological structure. 
% While other 3D assets, such as 3D meshes, leverage explicit topological structure to represent geometry. 
% Our future work will explore extending our method to diverse 3D asset formats and AIGC content.

%% file: sec/7_acknowledgement.tex
\section*{Acknowledgement}
This work was carried out at the Renjie Group, Hong Kong Baptist University. 
Renjie Group is supported by the National Natural Science Foundation of China under Grant No. 62302415, Guangdong Basic and Applied Basic Research Foundation under Grant No. 2022A1515110692, 2024A1515012822.